\documentclass{article}

\usepackage[final]{nips_2017}

\usepackage[utf8]{inputenc} 
\usepackage[T1]{fontenc}    
\usepackage{hyperref}       
\usepackage{url}            
\usepackage{booktabs}       
\usepackage{amsfonts}       
\usepackage{nicefrac}       
\usepackage{microtype}      
\usepackage{graphicx}       %

\usepackage{listings}
\title{Clinical Text Summarization with Syntax-Based Negation and Semantic Concept Identification} 

\author{
  Wei-Hung Weng \\
  \texttt{ckbjimmy@mit.edu} \\  
  \And
  Yu-An Chung \\
  \texttt{andyyuan@mit.edu} \\
  \And
  Schrasing Tong \\
  \texttt{st9@mit.edu}
}

\begin{document}

\maketitle
\section{Introduction}
In the era of clinical information explosion, a good strategy for clinical text summarization is helpful to improve the clinical workflow~\citep{charles2013adoption}.
The ideal summarization strategy can preserve important information in the informative but less organized, ill-structured clinical narrative texts.
Instead of using pure statistical learning approaches, which are difficult to interpret and explain, we utilized knowledge of computational linguistics with human experts-curated biomedical knowledge base to achieve the interpretable and meaningful clinical text summarization.
Our research objective is to use the biomedical ontology with semantic information, and take the advantage from the language hierarchical structure~\citep{hamburger1984acquisition}, the constituency tree, in order to identify the correct clinical concepts and the corresponding negation information, which is critical for summarizing clinical concepts from narrative text.  

We first used the classical natural language processing (NLP) approaches, such as sectioning, sentence and word tokenization, with Stanford CoreNLP~\citep{manning2014stanford}.
We then identifed clinical concepts in the narrative texts by several NLP components in the Apache clinical Text Analysis and Knowledge Extraction System (cTAKES)~\citep{savova2010mayo}, which include the clinical word-level morphological and semantic analyzers along with biomedical knowledge base, Unified Medical Language System (UMLS) and Semantic Network~\citep{bodenreider2004unified,mccray2003upper}.
For negation detection, we collected a list of negated terms (e.g. ``no'', ``rule out'', ``no evidence of'', ... etc.) and categorized them into different negation types.
Next, we detected negated concepts in the clinical sentences through sentence pruning, syntactic analysis and parsing using Apache OpenNLP and Stanford Tregex/Tsurgeon toolkits~\citep{levy2006tregex}.
Then, we combined the negation information of each identified concept, and constructed an curated itemized list of concepts by localizing the identified concepts in each document section.
Finally, we conducted case studies for error analysis and the direction of further improvement, and also evaluated the performance of the proposed method quantitatively.
The curated itemized list of clinically important concepts generated by our method may be the prototype for clinical effective communication.

In this study, we achieved the clinically acceptable performance for both negation detection and concept identification. 
The clinical concepts with common negated patterns can be identified and negated, yet we also found that the concepts inside fragmented sentences or sentences with medical abbreviations may not be correctly identified and negated.
The key challenge is that such sentences may not be parsed correctly by the statistical learning-based constituency tree parser that we relied on to obtain the correct syntactic patterns. 
Future directions of the project may focus on (1) designing more syntatcic rules for different sentence structure, especially for clinical fragmented sentences (2) investigating different constituency tree parsing models to fit the clinical content, and obtain correct constituency trees for downstream clinical syntax-based negation and clinical concept identification.

\section{Motivation}
Clinicians often face with the dilemma whether they should spend more time on seeing patients or go through clinical data carefully without missing any information.
Clinical information explosion, which results from the great adoption of electronic health record (EHR), aggravates this problem due to the exponential growth of the available clinical multimodal data~\citep{charles2013adoption}.

Among all modalities of clinical data, clinical narrative text is one of the most important sources for the appropriate patient care since it is not the raw physiological signals or raw imaging data but the information that is recorded by trained clinicians after human thinking and reasoning~\citep{weng2017medical}.
In other words, clinical narrative texts includes clinicians' thoughts and reasonings about patient's condition and treatment strategy. 
Therefore, reading medical records is essential for good clinical care, yet it is also a time-consuming task even for well-trained clinicians~\citep{oxentenko2010time}. 
This is because that clinicians are trained to use the specific documentation style---sentence structures with medical jargons and abbreviations, or semantically meaningful but syntactically fragmented sentences, to record patients' clinical history, disease progression and the results of examinations. 
Therefore, how to condense and summarize the document but keep the useful clinical concepts for the effective communication between clinicians becomes an important issue. 

Regarding text summarization task in the field of NLP, the state-of-the-art approach is deep learning algorithms, which utilizing the ability of stacked neural network layers that can approximate different mathematical functions to ``learn'' the semantics of words, sentences and documents.
For example, recent studies used neural networks with the sequence-to-sequence model or even reinforcement learning to demonstrate high performances on abstractive text summarization task in the general purpose~\citep{paulus2018deep,nallapati2016abstractive,rush2015neural}.

However, deep learning algorithms may not be the optimal choice for clinical corpus due to some reasons. 
First, clinical corpus is usually much smaller than general purpose corpus. 
For instance, the English Wikipedia has more than four million documents, yet the largest publicly available intensive care unit (ICU) clinical database, Medical Information Mart for Intensive Care (MIMIC-III) has less than 60,000 clinical summaries~\citep{johnson2016mimic}. 
For current deep learning approach, learning from small data is usually a big issue that is hard to overcome~\citep{lake2017building}.

Secondly, there is a huge biomedical knowledge base already existing~\citep{bodenreider2004unified}. 
This knowledge base has already integrated with many biomedical ontologies as well. 
Using biomedical concepts is helpful for text summarization since the texts can be normalized and itemized into the standardized terminologies in community consensus.
For example, different texts {\tt `cr', `Cr', `cre', `creat', creatinine} will be normalize to {\tt `Creatinine'} with appropriate ontology mapping.
In contrast, deep learning algorithms will take them as different, independent tokens.
There is also no feasible approaches to integrate this knowledge framework into machine learning and deep neural network architecture due to their fundamental differences.

Moreover, it is important to make the method much precise and interpretable rather than providing a transparent but hard-to-explained approximation function for clinical related tasks. 
Most clinical problems have a characteristics of near zero-tolerance of error~\citep{lipton2016mythos}. 
Thus, artificial intelligence method implementations in medicine usually require close inspection and evaluation by human clinical experts rather than just reporting few metrics.
For example, misidentifying patient with malignant tumor to without malignant tumor in the pathology report will result in devastating outcome.

For the automated clinical text summarization task, negation detection becomes one of the essential components since they play a critical role for preventing fatal errors which may result from computational models.
Current standard for the clinical negation is the regular expression-based method~\citep{chapman2001evaluation,chapman2013extending}.
The capability of using regular expression in linear string structure is interpretable but limited since human assembles language in a hierarchical geometrical structure rather than linear way~\citep{hamburger1984acquisition}. 
Researchers also approached the clinical negation problem from both the machine learning and linguistic-based methods rather than the simple regular expression approach~\citep{miller2017unsupervised,mehrabi2015deepen,gindl2008syntactical}.
Machine learning approach may be extraordinary to discover the patterns that human can not find. 
However, the model is hard to be interpreted by our knowledge of human language---which is essential for analyzing the linguistic structure.
These may be the reasons why the current clinical negation system can not be generalizable.

In this project, we therefore proposed a method and pipeline to integrate both the syntax-based negation detection and the semantic clinical concept identification for the clinical text summarization task for better clinical effective communication.

\section{Previous Work}
The key components of this project are clinical text summarization and clinical text negation detection.

\subsection{Clinical Text Summarization}
For clinical text summarization, current approaches are mainly based on extractive and indicative methods, which extract the original clinical text but not replace the original text to the clinical concept~\citep{pivovarov2016electronic,moen2016comparison,goldstein2016automated}. 

Most studies adopted machine learning algorithms to tackle the problem.
Pivovarov et al. performed machine learning algorithms, such as Latent Dirichlet allocation (LDA) and Bayesian networks to point out the important pieces in the original clinical document~\citep{pivovarov2016electronic}. 
Moen et al. extract key sentences using machine learning-based word space models~\citep{moen2016comparison}.
Goldstein and Shahar utilized the knowledge-based heuristics to extract the related sentences from the clinical text~\citep{goldstein2016automated}.

To our knowledge, there is no literature on clinical text summarization on the clinical concept level using itemized semantic concepts for clinical text summarization.
However, the itemized list instead of the paragraph is preferred since the former structure expression is easy to understand and well-structured.

\subsection{Clinical Negation Detection}
The current gold standard for the clinical negation detection is the regular expression-based method without considering language hierarchical structure~\citep{chapman2001evaluation}.
The standard algorithms, ConText and NegEx-ConText algorithms identify negated clinical concepts using regular expression by a list of negation trigger terms and linear token windows rather than the syntactic structure of sentence~\citep{chapman2001evaluation,chapman2013extending}.

Researchers also investigated the clinical negation problem through a linguistic approach.
Gindl et al. looked into the part-of-speech (POS) and the syntactic meaning of the negated term in the sentence~\citep{gindl2008syntactical}.
Mehrabi et al. adopted dependency parsing to improve the performance of regular expression based NegEx system~\citep{mehrabi2015deepen}.

These studies attepmted to look into the linguistic structure of sentence but did not consider the hierarchical structure of the sentence.

Others attempted to utilize the power of machine learning to discover the negation pattern in the different corpora as well. 
Miller et al. used unsupervised domain adaptation algorithm based on structural
correspondence learning (SCL) and boot-strapping~\citep{miller2017unsupervised}. 
Cruz et al. adopted the support vector machine algorithm to detect negation for sentiment analysis ~\citep{cruz2016machine}.
Fancellu et al. utilized the recurrent neural network architecture, bidirectional long short term memory (LSTM), to detect negation scope in non-biomedical corpora~\citep{fancellu2016neural}.
These machine learning methods usually yield promising performance but are hard to interpret and difficult to fix errors even if the error analysis was conducted.

In the study, we adopted the idea of the negation trigger term list and took the advantage of the syntactic structure of language for clinical negation detection.

\section{Methodology and Implementation}
Our proposed method allows us to summarize a clinical narrative document into an itemized clinical concept list by identifying clinical concepts with the corresponding negation information.
Both the sentence and fragment-level syntactic analysis, as well as the word-level morphological and semantic analysis for clinical named-entity recognition (NER) were used.

In the study, we first used the small set of sentences with negation, along with a list of curated negated trigger terms (196 negation terms and phrases) to design and implement a syntax-based negation detector (Section 4.2).
Then we adopted the clinical NLP pipeline and medical ontology to identify and extract biomedical concepts in the narrative texts through NER with knowledge base lookup (Section 4.3).
Next, the extracted concepts were filtered semantically.
Only the clinically relevant concepts were preserved to create the curated itemized list along with their negation information, which obtained from negation detector. 

\begin{figure}[ht]
  \centering
  \includegraphics[width=1\textwidth]{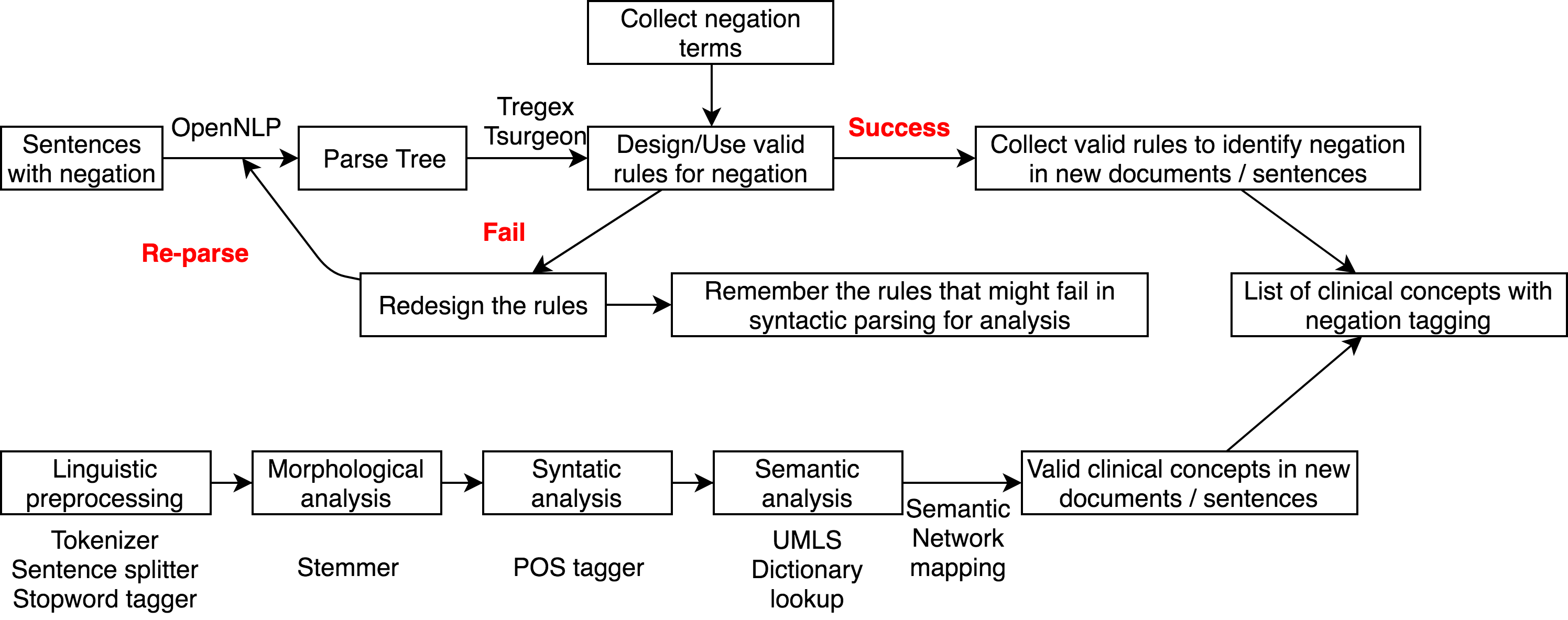}
  \caption{Workflow of syntax-based negation detection (upper) and semantic clinical concept identification (lower).}
  \label{fig:nlp}
\end{figure}

The full workflow is shown in Figure~\ref{fig:nlp}.
We also conducted error analysis for the clinical negation and see how to improve the performance of detecting negated concepts.

\subsection{Linguistic Preprocessing}
We applied Stanford CoreNLP toolkit for common linguistic preprocessing steps~\citep{manning2014stanford}, such as clinical document section and sentence fragmentation, word tokenization, stopword removal, before further natural language analysis tasks.

\subsection{Syntactic Analysis and Processing of Negation}
For detecting negation, we applied the Apache OpenNLP library, which is an open-source machine learning-based toolkit for NLP, to obtain constituency trees.
We also used Tregex and Tsurgeon, which are the tools for analyzing patterns of constituency trees and pattern matching via tree relationships and regular expression~\citep{levy2006tregex}.

The foundation of syntax-based negation detection is the collected negation trigger terms. 
We first collected negation trigger terms and formed a negation term list.
196 negation terms (both negation words and phrases) that are commonly used in clinical writing were collected from Multilingual NegEx Lexicon (MedInfo 2013, April 30th 2013, version 1)~\citep{chapman2013extending}, with manual modifications and additions.

We analyzed all collected negation terms and assigned them to the appropriate negation type regarding their negation location and negation phrase structure. 
For example, the negation term ``no'' belongs to ``pre-negation'' regarding the feature of location and {\tt ``NP''} of negation phrase type since ``no'' usually expresses in the pattern of ``no congestive heart failure'', ``no known allergy'', which the term is followed by a noun phrase.
Therefore, ``no'' has the characteristics of ``pre-negation'' and ``{\tt NP} type negation''.

\subsubsection{Negation Location}
The negation location is necessary for the meaningful sentence pruning before constituency tree parsing.
We designed five categories of the negation location~\citep{chapman2013extending}: 
Pre-negation (PREN), post-negation (POSN), pre-possible negation (PREP), post-possible negation (POSP), and pseudonegation (PSEU) which is not a true negation.
The motivation and further discussion of pesudonegation is in Section 5.

Table ~\ref{tab:neg} provides few examples of each type of negation location.
There are 83 pre-negation, 17 post-negation, 73 pre-possible negation, 8 post-possible negation, and 15 pseudonegation terms in our negation trigger term list. 

\begin{table}[htbp]
  \small
  \centering 
  \caption{Examples of commonly seen clinical negation terms in each negation location.}
  \label{tab:neg}
  \resizebox{\textwidth}{!}{%
  \begin{tabular}{|l|l|l|l|l|}\hline
    \textbf{Pre-negation} & \textbf{Post-negation} & \textbf{Pre-possible} & \textbf{Post-possible} & \textbf{Pseudonegation} \\ \hline
    no & free & ro & not ruled out & no increase \\
    no evidence & unlikely & r/o & did not rule out & no change \\ 
    without & was ruled out & rule out & not been ruled out & not cause \\ 
    cannot & is ruled out & rule out for & be ruled out & without difficulty \\
    unremarkable for & have been ruled out & may be ruled out for & is to be ruled out & not only \\ \hline
  \end{tabular}%
  }
\end{table}

\subsubsection{Negation Phrase Structure}
The goal of assigning the appropriate negation phrase structure is to explore general patterns of clinical negation, and design commands for the Tregex/Tsurgeon manipulation.
To be noticed, the negation phrase structure is not the phrase type of the negation term, but the type of interaction between the negation term and the negated concept.

We started from the simple rules and used real world clinical sentences to iteratively design our strategy and rules to identify more negation patterns, in order to achieve the generalizability.

The sentences are analyzed by human to detect the common patterns of acceptable negation meaning in the syntactic structures.
We mainly used bottom-up parsing with leftmost derivation in our manual investigation.
This characteristic is for appropriate Tregex/Tsurgeon manipulation of the sentences with different syntactic structures.

In general, there are eight categories. 
We found the following very common valid negation phrase structure patterns of clinical negation that are simple but can cover more than 80\% of constituency trees without overfitting --- NP negation (25), VP-A (VP anterior) negation (44), VP-P (VP posterior) negation (13), PP negation (70), ADJP-A (ADJP anterior) negation (1), ADJP-P (ADJP posterior) negation (8), ADVP-A (ADVP anterior) negation (16), and ADVP-P (ADVP posterior) negation (4).
Pseudonegation does not have the negation phrase structure type assignment since they are not valid negations at all.
These categories are all possible options for clinical negation (Table~\ref{tab:tregex}):

\begin{table}[htbp]
  \small
  \centering
  \caption{Common negation phrase structure patterns.}
  \label{tab:tregex}
  \resizebox{\textwidth}{!}{
  \begin{tabular}{|l|l|}\hline
    \textbf{Negation type} & \textbf{Example command in Tregex/Tsurgeon format} \\ \hline
    NP & {\tt NP=target << DT=neg <<, /no|without/ !> NP >> TOP=t} \\
     & {\tt delete neg} \\ \hline
    VP-A & {\tt VP=target << /VBZ|VBD|VB/=neg >> TOP=t} \\
     & {\tt delete neg} \\ \hline
    VP-P & {\tt VP=vp <<- /free|negative|absent|ruled|out|doubtful|unlikely|excluded|resolved|given/=neg \$ NP=head >> TOP=t >> S=s} \\
     & {\tt excise s head} \\ \hline
    PP & {\tt PP=head <<, IN=neg1 < NP=target >> TOP=t >> /S|NP|ADJP/=s \$ /JJ|NP/=neg2} \\ 
     & {\tt excise s target} \\ \hline
    ADJP-A & {\tt PP=head \$ /JJ|ADJP|NP/=neg <- NP=target >> TOP=t >> /S|NP/=s} \\
     & {\tt excise s target} \\ \hline
    ADJP-P & {\tt VP=vp <<- /free|negative|absent|ruled|out|doubtful|unlikely|excluded|resolved|given/=neg \$ NP=head >> TOP=t >> S=s} \\
     & {\tt excise s head} \\ \hline
    ADVP-A & {\tt VP=head \$ RB=neg <<, /VB*|MD/ >> TOP=t >> S=s} \\
     & {\tt excise s head} \\ \hline
    ADVP-P & {\tt VP=head \$ RB=neg <<, /VB*|MD/=be >> TOP=t >> S=s} \\
     & {\tt delete head,delete neg} \\ \hline
  \end{tabular}
  }
\end{table}

We also added few negation patterns to tackle some exceptions. For example, the sentence with {\tt SBAR}, with coordinate, or with some grammatical errors but commonly seen in clinical texts. 

\begin{figure}[ht]
  \centering
  \includegraphics[width=0.7\textwidth]{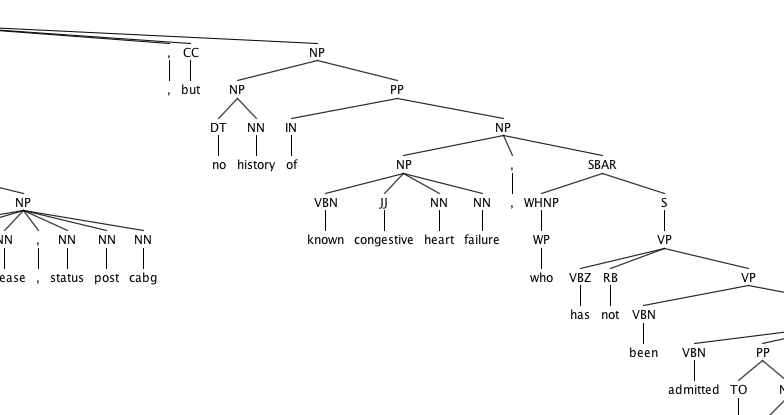}
  \caption{Examples of inappropriate subtree. In this case, the correct negated concept is ``congestive heart failure'' only. The {\tt SBAR} part should not be included for concept negation.}
  \label{fig:too-many}
\end{figure}

For excessive capturing of the negation span, we applied Tsurgeon to excise the subtrees.
For example, we can remove the {\tt SBAR} part of subtree under the {\tt NP} with negated concepts using the syntax: {\tt NP <<, no \& << SBAR=sbar | delete sbar} (Figure~\ref{fig:too-many}).
{\tt SBAR} is usually removable in most clinical sentences, especially after sentence pruning by localizing the negation term.

We listed few examples for some exceptions in Table~\ref{tab:except}.

\begin{table}[htbp]
  \small
  \centering
  \caption{Negation patterns for some exceptions.}
  \label{tab:except}
  \resizebox{\textwidth}{!}{%
  \begin{tabular}{|l|l|}\hline
    \textbf{Negation type} & \textbf{Example commands in Tregex/Tsurgeon format} \\ \hline
    Forced SBAR removal & {\tt SBAR=sbar} \\
     & {\tt delete sbar} \\ \hline
    NP without S node & {\tt NP=target <<, /DT|NN|RB/=neg <<, /no|without/ !> NP >> TOP=t} \\
    (for fragmented sentence) & {\tt delete neg} \\ \hline
  \end{tabular}%
  }
\end{table}

In our implementation, users can add new customized Tregex/Tsurgeon patterns and the corresponding conditions when they discover the new rules for negation patterns.

\subsubsection{Other Characteristics for Negation Types}
We also used the POS tagger model in NLTK ({\tt maxent\_treebanck\_pos\_tagger} model) to identify the POS of the first word in each negated trigger term.

We used this negation trigger term list with the additional designed features to detect negated concepts and their span in the following experiments.

\subsubsection{Sentence Pruning}
We extracted and pruned the sentences with negation patterns by identifying negation terms within the prepared negation term list.
We first identified whether the sentence includes the negation term.
If no, we then labeled them as non-negated sentence.
If yes, we then pruned the sentence based on the characteristics of negation location. 
For example, the sentence ``Left lower ext edema : U/S was performed , no evidence of dvt .'' will become ``no evidence of dvt .'' since ``no evidence of'' is a pre-negation negation term.
We assume that only the concepts after this negation term may be negated.
The purpose of the sentence pruning is to reduce the complexity of sentence parsing since the negation patterns of constituency trees become much complex when the sentence becomes longer.
For the longer sentence, the uncertainty of the statistical parser we used to obtain constituency trees may be higher.

\subsubsection{Sentence Parsing}
Next, the negated clinical sentences/fragments were imported into the OpenNLP pipeline and used its maximum entropy algorithm-based sentence parsing to obtain corresponding constituency trees.
Once the constituency tree is obtained, we utilized the syntactic parsing tools Tregex/Tsurgeon along with the corresponding negation phrase structure type to 
identify the negated subtrees and extracted them to match and negate the corresponding clinical concepts.

To extract the negated subtree by Tsurgeon, we applied top-down approach since that we have tried bottom-up method but the latter approach could not capture all possible negated concepts.
For example, the Tsurgeon syntax of extracting the subtree with the pattern of {\tt NP <<, no} is: {\tt NP=head <<, no >> TOP=t >> S=s | excise s head}.

we also experimented on using Stanford CoreNLP API for all tasks, from section and sentence tokenization to constituency tree parsing.
However, we shifted back to OpenNLP syntactic parsing with Tregex command line version due to the speed concern.

\subsubsection{An Example of Negation Detection}

The following is an artificial user case for the full process of negation detection ({\tt dvt}: deep vein thrombosis, {\tt U/S}: ultrasound):

\small
\begin{verbatim}
Clinical document
\end{verbatim}

$\downarrow$ Section and sentence tokenization by Stanford CoreNLP

\begin{verbatim}
`Left lower ext edema : U/S was performed , no evidence of dvt .'
\end{verbatim}

$\downarrow$ Obtaining the syntactic parse tree by OpenNLP 

\begin{verbatim}
(TOP (S (S (VP (VBD Left) (NP (JJR lower) (JJ ext) (NN edema))))
(: :) (S (S (NP (NNP U/S)) (VP (VBD was) (VP (VBN performed)))) (, ,)
(S (NP (NP (DT no) (NN evidence)) (PP (IN of) (NP (NN dvt)))))) (. .)))
\end{verbatim}

$\downarrow$ Extracting the subtree with the negated term and concept by Tregex with 

$\downarrow$ manual designed rules. In this case we used very simple rule: {\tt NP <<, no}.

\begin{verbatim}
(NP
  (NP (DT no) (NN evidence))
  (PP (IN of)
    (NP (NN dvt))))
\end{verbatim}



$\downarrow$ Extracting the negated clinical concept by regular expression

\begin{verbatim}
dvt
\end{verbatim}
\normalsize

This step is important for error analysis and improving the performance of the negation detection.
The modified version of the clinical negated term list is available in the GitHub repository~\citep{github-clneg}.

\subsection{Morphological, Syntactic and Semantic analysis}
We applied cTAKES for morphological and syntactic analysis for clinical concept NER~\citep{savova2010mayo}. 
cTAKES is an open-source NLP system for EHR free-text information extraction.
It is built on the basis of Unstructured Information Management Architecture (UIMA)~\citep{ferrucci2004uima}, which is an open source scalable and extensible platform for NLP. 

cTAKES has several analytical engines for processing narrative texts.
To identify correct and useful biomedical concepts, which are usually named entities in different clinical semantic categories, we adopted lexical normalizer, POS tagger and phrasal chunker to correctly perform NER task.

After the concept identification by cTAKES, we connected the concepts to the biomedical knowledge base UMLS Metathesaurus for semantic analysis and dictionary lookup~\citep{bodenreider2004unified}. 
UMLS Metathesaurus is a large, multi-purpose thesaurus with millions of biomedical and clinical related concepts and their relationships.  
We specifically used two clinical highly relevant ontologies in UMLS, Systematized Nomenclature of Medicine -- Clinical Terms (SNOMED-CT) and RxNorm, for concept semantic mapping.  
The standard terminology approach may be helpful for further analysis of using clinical narrative tasks for machine learning clustering, classification and prediction.

Then we further utilized the UMLS Semantic Network to filter clinically relevant UMLS concepts in clinical notes~\citep{mccray2003upper,mccray2001aggregating}. 
We adopted the trained clinician pre-defined semantic types and semantic groups for filtering (Table~\ref{tab:nec})~\citep{weng2017medical}.

\begin{table}[htbp]
  \centering 
  \caption{Designed clinical semantic categories and the corresponding examples.} 
  \begin{tabular}{|l|l|l|}\hline
    \textbf{Semantics} & \textbf{UMLS Semantic Types} & \textbf{Examples} \\ \hline
    Disorder/Syndrome & T047, T191 & hypertension, T2DM, afib, CHF \\ \hline 
    Symptom/Sign & T033, T040, T046, T048, & chest pain, cough, hypovolemic \\ 
     & T049, T184 &  \\ \hline 
    Medication & T116, T123, T126, T131 & epinephrine, ACEi, insulin, NS \\ \hline 
    Diagnostic procedure & T061 & CT scan, MRI of brain, PET, biopsy \\ \hline 
    Anatomy & T017, T024, T025 & left knee, right thyroid, segment VI \\ \hline
  \end{tabular}
  \label{tab:nec} 
\end{table}

\subsection{Tools}
We used use The Natural Language Toolkit (``nltk'') package and Stanford CoreNLP toolkit version 3.9.1 for text preprocessing and lexical normalization, such as word tokenization and stemming process~\citep{bird2004nltk}.  

We used Apache OpenNLP version 1.8.4, Tregex and Tsurgeon version 3.9.1 first by working with its GUI (TregexGUI) but will use the command line version later for batch processing~\citep{levy2006tregex} for negation syntax analysis.

The pipeline of morphological, syntactic and semantic analysis for clinical NER was integrated and implemented on python version 2.7.11 with cTAKES version 4.0.0.
UMLS Metathesaurus version 2016AB will be adopted for all clinical semantic mapping and filtering. 
All source codes and the instruction are available in the GitHub repository~\citep{github-clneg}.

\section{Testing}

\subsection{Data and Preprocessing}
For testing, clinical narrative texts were randomly selected from the MIMIC-III critical care database~\citep{johnson2016mimic}, which contains 58,976 ICU patients admitted to the Beth Israel Deaconess Medical Center (BIDMC), a large, tertiary medical center in Boston, Massachusetts, USA. 

We extracted narrative discharge summaries from the MIMIC-III database and randomly choose the result of three notes to demonstrate the capability of the proposed method. 
The three testing notes are not overlapped with the negated sentences used for developing rules (development set), as well as the sentences in evaluation set.

For evaluation, we used annotated negated sentences provided by Negex/ConText repository~\citep{github-negex}, which are also not overlapped with the development set. The detailed was shown in the section of evaluation.

We provide the development set and evaluation set in our code repository. 
However, testing notes are not available due to the terms and conditions in use of MIMIC-III narrative text data.
Therefore, we provided the slightly revised version of testing notes below for readers.

\subsection{Experiment}
In the testing phase, we demonstrated the result of summarizing three random clinical narrative notes in MIMIC-III.
We also performed the error analysis by discussing the issues we found in the real cases.
The error analysis gave us insights of the limitation and pitfall of our proposed method, especially the part of using machine learning-based statistical parser and in what conditions should we be cautious of using the tools based on statistical and machine learning approaches.
In the end, we provide conclusions of the investigation of the testing phase.

\subsubsection{Testing Case 1}
In the first testing case, we conducted the qualitative analysis of both concept identification and negation detection tasks.
We highlighted some key issues in our discussion of error analysis.
The original narrative text and the output of our proposed method are shown below.
Here we provided two sections of the note, ``history of present illness'' and ``brief hospital course'', since they are usually the most ill-structured sections in the clinical note.

\noindent\rule{\textwidth}{1pt}
\scriptsize
--- History of Present Illness ---
54 year old female with recent diagnosis of ulcerative colitis
on 6-mercaptopurine, prednisone 40-60 mg daily, who presents
with a new onset of headache and neck stiffness. The patient is
in distress, rigoring and has aphasia and only limited history
is obtained. She reports that she was awaken 1AM the morning of
 with a headache which she describes as bandlike. She
states that headaches are unusual for her. She denies photo- or
phonophobia. She did have neck stiffness. On arrival to the ED
at 5:33PM, she was afebrile with a temp of 96.5, however she
later spiked with temp to 104.4 (rectal), HR 91, BP 112/54, RR
24, O2 sat 100 \%. Head CT was done and relealved attenuation
within the subcortical white matter of the right medial frontal
lobe. LP was performed showing opening pressure 24 cm H2O WBC of
316, Protein 152, glucose 16.  She was given Vancomycin 1 gm IV,
Ceftriaxone 2 gm IV, Acyclovir 800 mg IV, Ambesone 183 IV,
Ampicillin 2 gm IV q 4, Morphine 2-4 mg Q 4-6, Tylenol 1 gm ,
Decadron 10 mg IV.  The patient was evaluated by Neuro in the
ED.
Of note, the patient was recently diagnosed with UC and was
started on 6MP and a prednisone taper along with steroid enemas
for UC treatment. She was on Bactrim in past but stopped taking
it for unclear reasons and unclear how long ago.
.
\begin{lstlisting}
--- History of Present Illness (itemized concepts) --- Diagnosis(+), Ulcerative Colitis(+), Headache(+), Neck stiffness(+), Distress(+), Rigor - Temperature-associated observation(+), Aphasia(+), Medical History(+), Headache(+), Headache(+), Phonophobia(-), Neck stiffness(+), Apyrexial(+), Pressure (finding)(+), Leukocytes(+), Proteins(+), Glucose(+), Vancomycin(+), Administration of enema(+), Therapeutic procedure(+)
\end{lstlisting}
\noindent\rule{\textwidth}{1pt}
--- Brief Hospital Course --- 54 woman on immunosuppressive therapy for UC (prednisone,
6MP) who presents with new onset HA, fever with bacterial
meningitis and gram positive rod bacteremia.

\#. Listeriosis - meningitis and bacteremia. Patient presented
with headache, nuchal rigidity, expressive aphasia, afebrile on
admission but temp to 104.4 in the ED, where she also started to
have rigors.  LP showed >300 WBC, poly predominant with 5%
monocytes, protein 152 glucose 16.  CSF gram stain showed gram
positive rods, blood culture grew gram positive rods, speciation
eventually grew listeria. Empiric treatment based on gram stain
was started: ampicillin and bactrim (to cover both nocardia and
question of PCP,  below), vanc and ceftriaxone as well pending
confirmation of gram stain and culture results.  Once
speciations was confirmed, a five day course of gentamicin was
started for synergy, and vancomycin and ceftriaxone d/c'd.
Bactrim was maintained on treatment dose for concern for PCP
 , when it was changed to prophylaxis dose.  Early on
admission, she developed hypotension that required levophed, but
was weaned off of pressors within the first couple of days of
admission with PRBCs (total of 4 units) and volume
resussitation.  Given bacteremia, TTE was done, no vegetations
or lesions noted. Head CT on admission showed right medial
frontal lobe likely infarct versus mass lesion, no hemorrhage.
Subsequent MRI confirmed infarct, unclear date, and EEG
consistent with meningitis.  Neurology was consulted, and the
patient was placed on dilantin for seizure prophylaxis given
meningoencephalitis. She spiked fevers to 101-102 over the first
several days of admission.  By , her neurological exam was
markedly improved, and by  her headache was gone, no
meningeal signs noted, although her baseline essential tremor
was slightly more severe. Surveillance blood cultures reamined
negative from  on.  Notably, she was transferred from ICU
to floor on , but noninvasive BP was read as 60/d, patient
mentating well, sent back to ICU. In the ICU, an arterial line
was placed, and consistently read 20-30 mmHg higher than
sphyngomanometer.  This discrepancy was of unclear etiology, but
persistent.  Patient maintained normal mentation, good urine
output, no tachycardia, and it was judged that, for some unclear
reason, the cuff pressures underestimated by 20-30 points.  On
, she was sent to the floor for further care and management.

\#. Bilateral lung opacities/hypoxia. Initial chest film read as
increased opacities bilaterally concerning for PCP (given
steroids and no PCP ) vs. bacterial pneumonia vs. pulmonary
edema. She had signifcant oxygen requirement, and her
respiratory distress led to her being placed on CPAP+PS.  The
origin of her significant hypoxia was originally thought to be
secondary to likely vascular leak from sepsis/CHF versus PCP. 
induced sputum was attempted, but was unsuccessful, and was not
repeated initally given her unstable respiratory status, and
susbsequent evaluation that likelihood of PCP was small.  She
responded well to lasix diuresis, with reduced O2 requirements.

\#. UC: She continued to receive her outpatient dose of
prednisone, which was changed on  to dexamethasone IV; her
outpatient 6-MP was held. After several days with no diarrhea,
it recurred on  soon after her diet had advanced. C.diff was
negative. She was made NPO, and fed via TPN for bowel rest.  On
, it was noted that she began passing BRBRP, her hematocrit
was noted to drop two points and pt was typed and crossed and
consent for blood transfusion.

\#. Anemia. On admission, she was found to be anemic. She
received PRBCs for anemia on admission and again  for mixed
venous sat <70\%.  She was found to have iron binding studies c/w
anemia of chronic disease.  Her HCT was followed closely, and
remained stable for the remainder of her admission.

\#. FEN: Her diet was advanced as tolerated, but she was made NPO
with TPN on  after she developed diarrhea, thought secondary
to continued UC activity.

\#. Prophylaxis: PPI. Hold SQ Hep, pneumoboots. Initially on
droplet precautions.

\#. Code status: FULL

\#. Communication: patient, her sister, brother, and mother

\#. Lines: peripheral IV x 2. left subclavian CC.  A-line. Eval
for PICC; once in place, can d/c central line, a-line.

\begin{lstlisting}
--- Brief Hospital Course (itemized concepts) --- Therapeutic immunosuppression(+), Fever(+), Meningitis, Bacterial(+), Rod Photoreceptors(+), Bacteremia(+), Listeriosis(+), Meningitis(+), Bacteremia(+), Headache(+), Nuchal Rigidity(+), Aphasia, Expressive(+), Apyrexial(+), Rigor - Temperature-associated observation(+), Leukocytes(+), Monocytes(+), Proteins(+), Glucose(+), Colony-Stimulating Factors(+), Rod Photoreceptors(+), Blood(+), Rod Photoreceptors(+), Therapeutic procedure(+), Vancomycin(+), Therapeutic procedure(+), Prophylactic treatment(+), Hypotension(+), Bacteremia(+), Lesion(-), Probable diagnosis(+), Infarction(+), Mass of body structure(+), Lesion(+), Hemorrhage(-), Infarction(+), Meningitis(+), Seizures(+), Prophylactic treatment(+), Meningoencephalitis(+), Fever(+), Several days(+), Headache(+), Physical findings(-), Essential Tremor(-), Grade 3 Severe Adverse Event(-), Blood(+), Discrepancy(+), Gait normal(+), Urine volume finding(+), Tachycardia(-), Pressure (finding)(-), Decreased translucency(+), Hypoxia(+), Abnormally opaque structure (morphologic abnormality)(+), Pneumonia, Bacterial(-), Pulmonary Edema(-), Respiratory distress(+), Continuous Positive Airway Pressure(+), Hypoxia(+), Probable diagnosis(+), Sepsis(+), Congestive heart failure(+), Unstable status(+), Several days(+), Diarrhea(-), NPO - Nothing by mouth(+), Parenteral Nutrition, Total(+), History of - blood transfusion(+), Anemia(+), Anemia(+), Iron(+), Anemia of chronic disease(+), NPO - Nothing by mouth(+), Parenteral Nutrition, Total(+), Diarrhea(+), Prophylactic treatment(+), Hepatoerythropoietic Porphyria(+), Respiratory secretion precautions(+), Colectomy(+), NPO - Nothing by mouth(+), Ostomy(+), Nausea(+), Disease(+), Rehabilitation therapy(+)
\end{lstlisting}
\normalsize
\noindent\rule{\textwidth}{1pt}

\paragraph{Concept Identification}
Initially, we found that using the original cTAKES NER component does not consider the span of concept and may obtain redundant concepts. 
For example, it will recognize ``ulcerative colitis'' as three concepts: ``ulcerative'', ``colitis'', and ``ulcerative colitis'' yet we just want to capture the longest concept ``ulcerative colitis''. 
Thus, we further implemented the script for detecting the longest extracted concept for these redundant concept identification.

Nevertheless, current concept identification approach may still miss some concepts. For example, we just identified ``phonophobia'' from ``photo- or phonophobia'' since ``photo-'' is not a usual clinical concept in our ontology.
The possible solution is to preserve the original text along with the mapped concepts to prevent information loss after the concept mapping step.

\paragraph{Negation Detection}
There are seven sentences in this note which contain negation terms. 

In the beginning, we missed the negated concepts of ``vegetation'', ``essential tremor'', but captured the affirmative concepts of ``pulmonary edema''. 
We found that the issue came from sentences with {\tt SBAR} structure that usually can not be handled by designed general rules.
For example, in ``her neurological exam was markedly improved , and by her headache was gone , \textbf{no} meningeal signs noted , although her baseline essential tremor was slightly more severe,'' the concept ``essential tremor'' in the {\tt SBAR} should not be negated, but the system indeed capture it.
Therefore, we designed the rule to handle the {\tt SBAR} condition by removing them is important for precise negation detection~\ref{tab:except}.

In the following testing cases we will mainly focus on the sentence with negation terms for negation detection since the issue of concept identification is similar across all inspected cases.

\subsubsection{Testing Note 2}

\paragraph{Concept Identification}
There is only one additional issue need to be pointed out regarding concept identification.
The clinical NER task using cTAKES with biomedical knowledge base and semantic information may miss capturing the concepts, or map to inappropriate concepts while ontology mapping due to ambiguity.  
For example, ``IVF'' can indicate ``intravenous fluid'' or ``fertilization in vitro''.
To capture correct clinical concepts, further investigation across sentences concerning semantics should be considered.
Cross-sentence semantics for identifying concepts may be achievable through the statistical approach, such as word embedding algorithms, based on co-occurrence of words.

\paragraph{Negation Detection}
There are 11 sentences in this note that contain negation trigger term. 
Most negation terms are correctly identified in this case.
For example, we can capture multiple negated concepts in the same sentence (``She otherwise denies any vomiting , rash , rhinorrhea , dysuria , cough , SOB or abdominal discomfort .''), which is not able to be identified if we use the Negex window size-based approach since they can set an arbitrary negated span (usually five words) before or after the negation trigger term. 

In this testing case, we can still remove the redundant part of the sentence, e.g. {\tt SBAR} part, to reduce the unnecessary negation span.

However, we also found that some medical terms can not be identified by OpenNLP parser and therefore they will be ignored during the construction of constituency trees. 
For example, the negated medication concept ``Neulasta'' was not recognized by statistical parser and therefore was not be negated in the end. 
The reason may due to the fact that such clinical terms are not shown in the original training dataset for OpenNLP parser.
The constituency tree of this case is as following:

\footnotesize
{\tt
--- Original --- Most likely this current episode of neutropenia is due to \textbf{the fact that Neulasta was not given during this cycle} of chemo per pt 's request , however due to the rapid rise in her WBC count myelosuppression from sepsis was also a possibility .

--- Tree --- (TOP (S (S (NP (NP (ADJP (RBS Most) (JJ likely)) (DT this) (JJ current) (NN episode)) (PP (IN of) (NP (NN neutropenia)))) (VP (VBZ is) (ADJP (JJ due) (PP (TO to) (NP (DT the) (NN fact) (SBAR (IN that) (VBN given) (PP (IN during) (NP (NP (DT this) (NN cycle)) (PP (IN of) (NP (NP (NP (NN chemo)) (PP (IN per) (NP (NP (NN pt) (POS 's)) (NN request)))) (, ,) (ADJP (RB however) (JJ due) (PP (TO to) (NP (NP (DT the) (JJ rapid) (NN rise)) (PP (IN in) (NP (PRP\$ her) (NNP WBC) (NN count) (NN myelosuppression))) (PP (IN from) (NP (NN sepsis)))))))))))))))) (VP (VBD was) (ADVP (RB also)) (NP (DT a) (NN possibility))) (. .)))
}
\normalsize

One can see that ``Neulasta'' was gone in the tree.
A potential solution might be replacing professional clinical terms (especially the name of medication or clinical procedure) to some special token, such as ``[MEDICATION]'', and record the original name along with their location in other place. 
Once the ``[MEDICATION]'' is captured and negated, we can replace them back to the original name of medication.

Otherwise, the negation detection did well in this case. 
The input and output of the proposed pipeline are shown as following.

\noindent\rule{\textwidth}{1pt}
\scriptsize
--- History of Present Illness --- 
... She stated that for the past
two days she has noticed an increasing amount of stool output in
her ostomy bag but denies abdominal discomfort or blood in her
stool. ... She otherwise denies any vomiting, rash,
rhinorrhea, dysuria, cough, SOB or abdominal discomfort. She
denies any recent travel or sick contacts as well. ... His CXR did not
show definitive source of infection either. ... While in the ED she developed
hypotension not responding to IVF boluses, the pt denied CVL
placement and required the initiation of phenylepherine
peripherially in order to maintain SBPs in the 90s-100s. She did
not have a change in her mentation during these episodes of
hypotension. ...

\begin{lstlisting}
--- History of Present Illness (itemized concepts) --- Ostomy(+), Abdominal discomfort(-), Blood(-), Vomiting(-), Exanthema(-), Rhinorrhea(-), Dysuria(-), Coughing(-), Dyspnea(-), Abdominal discomfort(-), Illness (finding)(-), Source of infection(-), Hypotension(+), Fertilization in Vitro(+), Hypotension(-)
\end{lstlisting}
\noindent\rule{\textwidth}{1pt}
--- Brief Hospital Course --- 
...
\#Neutropenic Fever- On presentation the pt's PMN count was 21
most likely from her most recent chemotherapy cycle and lack of
Neulasta use. ...  When cdiff returned negative, flagyl was
discontinued. Blood cultures were sent and a U/A was not
concerning for infection. We also sent off galactomannan antigen
and beta-D-glucan labs initially as part of her neutropenic
fever workup which were negative. The following day after
admission her WBC rose significantly and she no longer was
neutropenic. ... She was continued on
Vanc/Cefepime until afebrile and with ANC>1000 for greater than
48 hours, after which she was switched to PO levofloxacin to
complete an 8 day total course for community acquired pneumonia.
...
\# Hypotension- In the  pt's SBP dropped to 70s, not responding
to IVF boluses. ... She was not administered
any medications recently that could be accounting for her
hypotension. ... Most likely this current episode of
neutropenia is due to the fact that Neulasta was not given
during this cycle of chemo per pt's request, however due to the
rapid rise in her WBC count myelosuppression from sepsis was
also a possibility.
...
\begin{lstlisting}
--- Brief Hospital Course (itemized concepts) --- Blood(+), Communicable Diseases(-), Leukocytes(+), Hypotension(+), Fertilization in Vitro(-), Hypotension(+), Probable diagnosis(-), Neutropenia(+), Neulasta(+), Chemotherapy(+), Leukocytes(+), Myelosuppression(+), Sepsis(+)
\end{lstlisting}
\normalsize
\noindent\rule{\textwidth}{1pt}

\subsubsection{Testing Note 3}
Again, we only focus on the sentences with negated terms for negation detection. 
There are 12 sentences in this note which contain negated terms.
In general, the negation detection can find the negated concepts correctly in most cases.

The issue raises in this case is that sometimes using only Tregex/Tsurgeon can not remove the negation term due to the structure of constituency tree.
For example, in the following case the negation term ``negative for'' is splitted and hard to be removed by Tregex/Tsurgeon.

\footnotesize
{\tt
--- Original --- Abdominal CT was negative for an abscess or fluid collection surrounding the pump .

--- Subtree --- (TOP (NP (NP (NN negative)) (PP (IN for) (NP (NP (DT an) (NN abscess) (CC or) (NN fluid) (NN collection)) (VP (VBG surrounding) (NP (DT the) (NN pump))))) (. .)))
}
\normalsize

The possible solution again is simply use regular expression to remove the negation term.
However, we did not specifically implement this part since the negation terms will be removed anyway after the clinical concept identification and mapping.

For future refinement, we may experiment on whether using constituency tree structure for negating concepts, and using linear regular expression for removing negation trigger terms will be helpful.

\noindent\rule{\textwidth}{1pt}
\scriptsize
--- History of Present Illness --- 
... CXR showed
cardiomegaly but no infiltrate. Abdominal CT was negative for an
abscess or fluid collection surrounding the pump. ...
\begin{lstlisting}
--- History of Present Illness (itemized concepts) --- Cardiomegaly(+), Infiltration(-), Abscess(-)
\end{lstlisting}
\noindent\rule{\textwidth}{1pt}
--- Brief Hospital Course --- 
... Pt went to surgery to have pump removed and the
operation appeared to be without complication. ... A TTE ruled out endocarditis in
the setting of group B strep bacteremia. ... Surgery was consulted and they
diagnosed a seroma and recommended conservative management given
that it had no signs of infection.
...  The culture returned positive for MRSA and gram
negative rods (not pseudomonas).  ...
Once the GNR sensitivities showed that it was not pseudomonas,
ceftaz was stopped.  ... She was
without signs of baclofen withdrawal (i.e. incr HR, temp, BP,
seizures) once pump was removed. ...  An EEG on HD \#3 was consistent with
severe encephalopathy and an EEG on HD \#6 was consistent with
meningoencephalitis with no evidence of seizures.  An MRI was
finally done on HD \#10 and showed mild communicating
hydrocephalus, no evidence of cavernous thrombosis or stroke.
...
After a supposed seizure at OSH pt was mechanically ventilated
thru her trach site b/c no breath sounds were appreciated. (At
baseline, pt has respiratory weakeness 2/2 to multiple sclerosis
but does not require mechical ventilation. ...
\begin{lstlisting}
--- Brief Hospital Course (itemized concepts) --- Complication(-), Endocarditis(-), Group B streptococcal pneumonia(-), Streptococcal Infections(-), Bacteremia(-), Seroma(+), Conservative Treatment(+), Physical findings(+), Communicable Diseases(-), MRSA - Methicillin resistant Staphylococcus aureus infection(+), Rod Photoreceptors(+), Antimicrobial susceptibility(+), Physical findings(-), Baclofen(-), Seizures(-), Grade 3 Severe Adverse Event(+), Encephalopathies(+), Meningoencephalitis(+), Seizures(-), Mild Adverse Event(+), Communicating Hydrocephalus(+), Thrombosis(-), Cerebrovascular accident(-), Seizures(+), Multiple Sclerosis(+)
\end{lstlisting}
\normalsize
\noindent\rule{\textwidth}{1pt}

\paragraph{Other Interesting Patterns}
We found that the pre-negation term is commonly seen in the middle of noun phrase. 
For example,``neurologic - \textbf{no} numbness, tingling, or weakness.'' ({\tt (NP (NP (NP (NN neurologic)) (: -) (NP (DT no) (NN numbness)) (, ,) (NP (NN tingling))) (, ,) (CC or) (NP (NN weakness)) (. .))}), which commonly happens in examination sections. 
Our sentence pruning strategy is helpful to remove the excessive negation in such cases. 
However, this may also result in the trade-off of the appearance of fragmented sentence and the capability of statistical parser.

We also found the existence of double negation in the clinical texts.
The commonly seen double negation problem is due to the negation of already negated words. 
For example, if the negated word is ``no abnormal''---``abnormal'' has already contained the semantics of negation, then it is NOT negated after adding ``no'' in front of it, and we should labeled them as ``pseudonegation''. 
We designed the specific negation type for commonly seen clinical pseudonegation terms to prevent such false negation.

\subsubsection{Conclusion of the Testing Phase}
From the above analysis of real testing cases, we realize that proposed approach gives us precise summarized information but still have a space to improve the performance.

\paragraph{Concept Identification}
Using the semantic concept identification approach has the advantage of standardization. 
The extracted concepts are standardized by biomedical knowledge base and ontology, so the issues of writing style and English grammar can be ignored as much as possible.
The standardized concepts can also be processed and compared afterward for large scale clinical text analysis as well.

However, for the concept identification task there are still some parts that we can improve beyond the span of concept that we have solved.
There are still some concepts that are not correctly captured.
For example, some medications are missing in the summarization. 
We may need to relax the restriction of semantic filtering by adding more semantic types. 
However, we need to keep in mind the trade-off between the amount of necessary information and usefulness of the information.
Presenting both the original text and the identified UMLS concept, as well as using word embedding approach to enhance the semantics between sentence may also be helpful for improvement.

\paragraph{Negation Detection}
For the negation detection task, our approach worked well with acceptable performance judge by the clinician since it can capture most of the negated concepts if they are correctly identified by semantic concept identification. (Some of the correctly negated concepts are removed during concept identification.)

We can handle common double negation cases by introducing the family of pseudonegation terms. 
However, different from general negation cases, there are still some very special clinical negations that we are not yet able to capture well. 
For example, ``afebrile'' means ``no fever'' (test case 1). 
We need to explore other strategy to identify those concepts contain the semantics of negation WITHIN the concept, which are not able to be correctly negated by using the negation trigger terms list approach.

In the sentence level, we pruned the whole sentence into the possible negated portion using negation trigger term list and their features of pre- or post-negation
The advantage of this approach is that we can minimize the error parsing that usually happens in OpenNLP or any statistical parsing tools. 
However, the following issue of this approach is that the sentence may be fragmented. 
The fragmented issue is also a common problem in raw clinical texts. 
For example, ``No acute distress. '' or ``Without any evidence of significant anemia.'' 

We need to consider more about the syntactic structure with complicated condition such as {\tt SBAR} or fragmented sentences, which may be parsed incorrectly by OpenNLP parser.
String pattern matching instead of constituency tree parsing is inevitable to correctly identify these cases.

Initially, we attempted to solve the problem through rewriting the sentences. 
However, it is not feasible to generalize patterns for rewriting the clinical negation sentences. 
Thus, we instead approached the problem through adding more rules to deal with exceptions, which may be much helpful to improve the performance of negation detection even if we face the new text corpus in the future.

\paragraph{Tools}
There are some issues raised from the tools we adopted in the project. 

\begin{figure}[ht]
  \centering
  \includegraphics[width=0.3\textwidth]{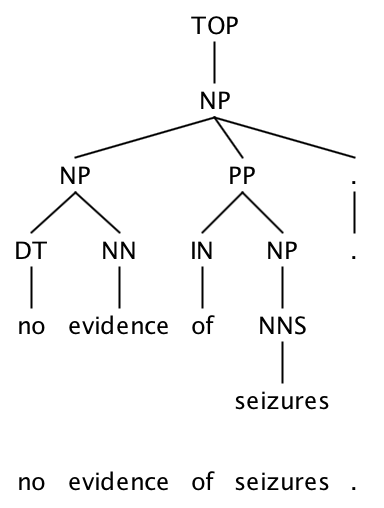}
  \caption{OpenNLP parsing for {\tt PP} type negation, ``no evidence of''.}
  \label{fig:noev}
\end{figure}

\begin{enumerate}
  \item Tregex can only operate one node at once. This limited us to do only one-step manipulation in one command. Thus, multiple steps processing is required for complicated cases.
  \item Tregex can not handle multiple matches. 
  However, since the goal of our task is to detect the span of negation, it is okay to cover whole span without removing all negated terms. 
  For example, once we detected the first ``no'' in ``She otherwise no vomiting, no rash, no rhinorrhea, no dysuria, no cough, no SOB and no abdominal discomfort.'', then we can just mark all concepts after this negation term without specifically removing other ``no''---which means that single match is useful enough for us to detect the negated concepts.
  \item The maximum entropy-based OpenNLP parsing algorithm can not always parse sentences in a correct way. 
  \item There are some issues for {\tt ADJP} and {\tt PP} negated types in OpenNLP parsing. 
  ``no evidence of'' and ``negative for'' are usually splitted by OpenNLP. 
  We need to remove the negation term arbitrarily through linear string pattern matching if needed.
  However, for negating the concepts, this is a minor issue since we will not preserve those negation terms after concept identification through mapping biomedical ontology. 
  For example, OpenNLP parses ``no evidence of seizures'' to the tree like Figure~\ref{fig:noev}.
  Tregex will only crop out ``no evidence'' and leave ``of seizures'', which is not the exact result we want.
  Yet we will only obtain the correct negated concept ``seizures'' after ontology mapping.
\end{enumerate}

\paragraph{Result Presentation}
For visualizing the result, a better approach to present our result to make our itemized list be easier to interpret is needed.
For instance, building a hierarchical structured tree for detailed items as we see in the file system of computers. (e.g. Brief Hospital Course >> Congestive Heart Failure >> Treatment >> several concepts about medications)



\section{Evaluation}
We evaluated the performance of negation detection using 324 negated concepts in 173 clinical negated sentences.
For evaluation, we reported the accuracy of negation detection. 
We briefly discussed the result and did analyzed the error cases in two different approaches since the human explanation is required for potential clinical use---we need to make the proposed method be explainable and interpretable. 

We provide the evaluation set in our code repository under {\tt test\_ready.txt}, which is derived from {\tt Annotation-1-120.txt} (from the Negex/ConText algorithm) with our human labeling.

\subsection{Evaluating Negation Detection}
For the negation detection, the clinician was asked to examined the output of all 173 negated sentences, and labeled the negated clinical concepts inside the sentences. 
Then, the clinician evaluated whether the method captured or missed the negated concepts manually. 
We also ran the baseline regular expression-based Negex algorithm for comparison.

Then we computed the accuracy of negation detection between our proposed approach and Negex approach.

The performance of negation detection is shown in Table~\ref{tab:perf_neg}.

\begin{table}[htbp]
  \centering 
  \caption{Performance of negation detection.}
  \label{tab:perf_neg} 
  \begin{tabular}{|l|l|l|l|l|}\hline
    \textbf{Method} & \textbf{Accuracy} \\ \hline
    Syntax-based (proposed approach) & 0.929 (301/324) \\ \hline
    Regular expression-based (implemented Negex algorithm) & 0.889 (288/324) \\ \hline
  \end{tabular}
\end{table}

We hypothesized that syntax-based negation detection works better than baseline regular expression-based detection since the latter approach considers only the language linear structure but not the hierarchical structure. 
The regular expression-based approach uses the window size and punctuation to determine the span of negation, which is intuitive but may be limited to capture the negation in complex or long sentences.


However, the syntax-based approach still faces problems due to the limitation of the statistical parser that generates constituency trees---in our case, the maximum entropy-based Apache OpenNLP parser.
We found that some rules we designed to prune and extract the constituency subtrees are not optimized. 
For example, in the following examples (we demonstrated in the format of our intermediate result during processing):

\footnotesize
\begin{verbatim}
sent: 77
original: 04 ) normal saline contrast injection without 
evidence of right to left intracardiac shunt .	 [NEGATED]
neg part: without evidence of right to left intracardiac shunt .
negated term: without evidence of
--- tregex/tsurgeon with negated type: PP
constituency tree: (TOP (PP (IN without) (NP (NN right) (SBAR (S (VP (TO to) (VP (VB left) 
(NP (JJ intracardiac) (NN shunt))))))) (. .)))
--- NP without S node
constituency tree: (TOP (PP (IN without) (NP (NP (NP (NN evidence)) (PP (IN of) (
NP (NN right)))) (SBAR (S (VP (TO to) (VP (VB left) (NP (JJ intracardiac)
(NN shunt))))))) (. .)))
--- forced remove SBAR
constituency tree: (TOP (PP (IN without) (NP (NP (NP (NN evidence)) (PP (IN of) (NP 
(NN right))))) (. .)))
--- remove first token f if f in negated list or stopword list
>> evidence of right .
\end{verbatim}
\normalsize

OpenNLP parsed this fragmented sentence in a strange way and split the concepts that should be negated: ``right to left intracardiac shunt'' to ``evidence of right'' ({\tt NP}) and ``to left intracardiac shunt'' ({\tt SBAR}), which is not correct.
Therefore we did not obtain the correct negated concepts in this case due to the incorrect constituency tree.

However, this is an issue for all clinical narrative text analysis tasks since clinical texts usually have a characteristics of fragmentation, abbreviation, as well as using incorrect grammars for clinical convenience and time saving.

To further improve the performance of syntax-based negation detection, the best approach to improve the statistical model is to collect the human annotation of POS tagging and constituency tree tagging for fragmented sentences and clinical abbreviation. 
However, this may not be a feasible approach since it is a labor (clinical expert)-intensive task.

The alternative way is to investigate other potential statistical models to obtain constituency trees. 
Using different machine learning models may decrease the interpretability but may be helpful to boost the performance.
For example, using recurrent neural network (RNN)-based sequential models~\citep{sutskever2014sequence}, or even the hierarchical model such as TreeLSTM~\citep{tai2015improved} that might be possible to preserve the semantics in the hierarchy.


\subsection{Evaluating Concept Identification}
For the concept identification, we also asked the clinician to label the meaningful clinical concepts in the sentences of evaluation set by his expertise and examined the result of automated extracted clinical concepts.
Then, we computed the accuracy, precision, recall and F1 score to investigate whether cTAKES clinical NER with UMLS Metathesaurus knowledge base can capture meaningful clinical concepts with/without using Semantic Network filtering for only clinically relevant concepts. 

The performance of concept identification is shown in Table~\ref{tab:perf_con}. 

\begin{table}[htbp]
  \centering 
  \caption{Performance of concept identification using cTAKES clinical NER and UMLS knowledge base, with or without Semantic Network filtering for clinical concepts.}
  \label{tab:perf_con} 
  \begin{tabular}{|l|l|l|l|l|}\hline
    \textbf{Score} & \textbf{with semantic filtering} & \textbf{without semantic filtering} \\ \hline
    Accuracy  & 0.719 & 0.397 \\ \hline
    Precision & 0.843 & 0.432 \\ \hline
    Recall    & 0.830 & 0.830 \\ \hline
    F1-score  & 0.836 & 0.568 \\ \hline
  \end{tabular}
\end{table}

Utilizing the strategy of UMLS semantic filtering can significantly reduce the case number of false positive and therefore yields much precise summarized information.
This is not surprised since the full set of UMLS Metathesaurus contains many redundant concepts with semantic types that are not related to clinical work. 
For example, ``Amino Acid, Peptide, or Protein'', ``Embryonic Structure'', ``Regulation or Law'', ``Experimental Model of Disease'', ``Gene or Genome'', ... etc. 
These concepts might be useful for biomedical literature analysis, but not for clinical documents.

However, there are still two main issues of using cTAKES and UMLS Metathesaurus mapping.  
There are issues of ambiguity and granularity in concept identification. 
For example, ``posterior neck pain'' versus ``neck pain'', ``drug allergy'' versus ``allergy'', ``tongue elevation'' versus ``elevation'', ``occult blood'' versus ``blood''.
cTAKES clinical NER sometimes gives us the concept with higher granularity but sometimes not. 
We may want to map to concepts using the largest word span to minimize the uncertainty of the ambiguity and granularity issue.

Occasionally, UMLS Metathesaurus has both affirmative and negated concept identifiers for some clinical concepts. 
To prevent from incorrect negation, we may need to improve the negation detection to ensure that we will not leave the negation terms in the input for concept identification.



\section{Conclusion}
In this study, we demonstrated that the proposed method of combining the syntax-based negation detection and semantic concept identification provides us the ability for clinical narrative summarization.

The key contributions of this project are:
\begin{enumerate}
 \item Designing and developing a strategy and rules of classifying negated trigger terms for better negation detection
 \item Utilizing the constituency tree to identify the negation pattern syntactically and apply different rules to identify the neagted concepts
 \item Identifying the clinically relevant concepts using biomedical knowledge base with semantic information along with syntax-based negation information
\end{enumerate}
We also realized that the language linear structure is efficient but not enough to capture the correct syntax. 
Hierarchical nature of human language should not be ignored for such natural language tasks.
Having the expert knowledge of clinical semantic information also allow us to obtain the better concept identification and summarization.

For future investigation, we plan to work on improving and customizing the constituency parser for clinical narrative texts. 
We also plan to design more rules for fragmented sentences and abbreviations that are usually seen in clinical content. 
Based on the human knowledge of clinical language, we also plan to improve the statistical learning part of sentence parsing with different hierarchical learning models, such as TreeLSTM.
We hope that the proposed method can be a prototype of the clinical text summarization tool for effective clinical communication.

\bibliography{mybib}
\bibliographystyle{abbrvnat}

\pagebreak

\section*{Appendix A~~~~Instruction of Using the Implementation}
We demonstrate the instruction of using the codes in this section ({\tt README.md}). 
Please refer to~\url{https://github.com/ckbjimmy/clneg} to download and use the codes.
The complete implementation with all dependencies configured is also available at~\url{https://www.dropbox.com/s/zw0uod64nt5wsf0/clneg.zip?dl=0}.
However, we can not provide our own UMLS licensing and access for public use.
Please request the access through UMLS Terminology Services website~(\url{https://uts.nlm.nih.gov/home.html}).
After getting the access, you can simply follow the instruction to run the experiment.

\normalsize
\begin{lstlisting}
# Clinical Text Summarization Tool with Syntax-based Negation and Semantic Concept Identification


## Introduction
we utilized the power of computational linguistics with human experts-curated knowledge base for identifying clinical concepts with their corresponding negation information in the clinical narrative texts.

We used the medical knowledge base UMLS along with Semantic Network, and take the advantage from the language hierarchical structure, the constituency tree, in order to identify the clinically relevant concepts and the negation information, which is extremely important for summarization.

In this project, we used Stanford CoreNLP, Apache clinical Text Analysis and Knowledge Extraction System (cTAKES), Unified Medical Language System (UMLS) and Semantic Network, to identify clinical concepts in the narrative texts.
We also performed the negation detection in the clinical sentences through sentence pruning, syntactic analysis and parsing using Apache OpenNLP and Stanford Tregex/Tsurgeon.
Then, we constructed itemized lists of clinically important concepts using the information generated from concept identification and negation.


## Dependencies
We provide `setup.sh` to download, install and configure most of the dependencies. 
The process is about 5 minutes.

- [Stanford CoreNLP 3.9.1](http://nlp.stanford.edu/software/stanford-corenlp-full-2018-02-27.zip)
- [Stanford Tregex/Tsurgeon 3.9.1](https://nlp.stanford.edu/software/stanford-tregex-2018-02-27.zip)
- [Apache OpenNLP](https://www.apache.org/dyn/closer.cgi/opennlp/opennlp-1.8.4/apache-opennlp-1.8.4-bin.tar.gz)
- [Apache cTAKES](http://ctakes.apache.org/)

However, you still need to request the access to UMLS by yourself to ensure that you can run semantic concept identification.

- [Access to NLM UMLS Metathesaurus/Ontology](https://www.nlm.nih.gov/databases/umls.html)
- The complete implementation with all dependencies configured is also available at `https://www.dropbox.com/s/zw0uod64nt5wsf0/clneg.zip?dl=0` (without UMLS account and password).

## Instruction
1. Request the UMLS access (this step will require few days for NIH/NLM to inspect your access application)
2. Run `sh setup.sh` in the first time
3. Add your UMLS account/password to `./src/ctakes/bin/pipeline.sh` after `-Dctakes.umlsuser=` and `-Dctakes.umlspw=`
4. Run `sh run_corenlp.sh` and `sh run_opennlp.sh` to initialize Stanford CoreNLP server and Apache OpenNLP server (make sure they are running in the background at port 9000 and 8080, respectively)
5. Open the other terminal and run `python main.py [file_path]`

- `../data/dev.txt` for development set
- `../data/test_ready.txt` for evaluation set
- `../data/1.txt` for testing note (as well as `2.txt`, `3.txt`) (The notes are no longer be available here. Please request the DUA of MIMIC-III database for using the notes)

6. We modularized some mutable components into files

- To design and add more rules for Tregex/Tsurgeon, please edit `tree_rules.py`
- To add more negation terms, please edit `./data/neg_list_complete.txt`, which is the modified version of the original `multilingual_lexicon-en-de-fr-sv.csv` open sources with our annotations
- To change the clinical semantic types for filtering, please edit `concept_extraction.py`

7. Baseline can be obatined by running `negex.py`
8. Please check the terminal screen for sentence parsing, and check the `./data/final_output`
9. For the process of development, please check the jupyter notebook `nlp_dev.ipynb` under `src` folder

\end{lstlisting}

\pagebreak

\section*{Appendix B~~~~Output Example}
We demonstrate the command line prompt and output of the implementation using the sentences in development set.  
The last line in the output demonstrates that the method detected the negated concepts (present in the negative sign) in the sentences.
For example, {\tt cancer(+)} means that the cancer exists, {\tt cancer(-)} means that it is instead negated.
Please refer to Appendix Section A or~\url{https://github.com/ckbjimmy/clneg} for the instruction of using the codes.

\scriptsize
\begin{lstlisting}
dhcp-18-111-118-162:src weng$ python main.py ../data/dev.txt
Parsing CPE Descriptor
Instantiating CPE
log4j: reset attribute= "false".
log4j: Threshold ="null".
log4j: Retreiving an instance of org.apache.log4j.Logger.
log4j: Setting [ProgressAppender] additivity to [false].
log4j: Level value for ProgressAppender is  [INFO].
log4j: ProgressAppender level set to INFO
log4j: Class name: [org.apache.log4j.ConsoleAppender]
log4j: Parsing layout of class: "org.apache.log4j.PatternLayout"
log4j: Setting property [conversionPattern] to [%m].
log4j: Adding appender named [noEolAppender] to category [ProgressAppender].
log4j: Retreiving an instance of org.apache.log4j.Logger.
log4j: Setting [ProgressDone] additivity to [false].
log4j: Level value for ProgressDone is  [INFO].
log4j: ProgressDone level set to INFO
log4j: Class name: [org.apache.log4j.ConsoleAppender]
log4j: Parsing layout of class: "org.apache.log4j.PatternLayout"
log4j: Setting property [conversionPattern] to [%m%n].
log4j: Adding appender named [eolAppender] to category [ProgressDone].
log4j: Level value for root is  [INFO].
log4j: root level set to INFO
log4j: Class name: [org.apache.log4j.ConsoleAppender]
log4j: Parsing layout of class: "org.apache.log4j.PatternLayout"
log4j: Setting property [conversionPattern] to [%d{dd MMM yyyy HH:mm:ss} %5p %c{1} - %m%n].
log4j: Adding appender named [consoleAppender] to category [root].
14 May 2018 13:10:54  INFO Chunker - Chunker model file: org/apache/ctakes/chunker/models/chunker-model.zip
14 May 2018 13:10:56  INFO TokenizerAnnotatorPTB - Initializing org.apache.ctakes.core.ae.TokenizerAnnotatorPTB
14 May 2018 13:10:56  INFO ContextDependentTokenizerAnnotator - Finite state machines loaded.
14 May 2018 13:10:56  INFO AbstractJCasTermAnnotator - Using dictionary lookup window type: org.apache.ctakes.typesystem.type.textspan.Sentence
14 May 2018 13:10:56  INFO AbstractJCasTermAnnotator - Exclusion tagset loaded: CC CD DT EX IN LS MD PDT POS PP PP$ PRP PRP$ RP TO VB VBD VBG VBN VBP VBZ WDT WP WPS WRB
14 May 2018 13:10:56  INFO AbstractJCasTermAnnotator - Using minimum term text span: 3
14 May 2018 13:10:56  INFO AbstractJCasTermAnnotator - Using Dictionary Descriptor: org/apache/ctakes/dictionary/lookup/fast/sno_rx_16ab.xml
14 May 2018 13:10:56  INFO DictionaryDescriptorParser - Parsing dictionary specifications:
14 May 2018 13:10:56  INFO UmlsUserApprover - Checking UMLS Account at https://uts-ws.nlm.nih.gov/restful/isValidUMLSUser for user ckbjimmy2:
..14 May 2018 13:10:57  INFO UmlsUserApprover -   UMLS Account at https://uts-ws.nlm.nih.gov/restful/isValidUMLSUser for user ckbjimmy2 has been validated

14 May 2018 13:10:57  INFO JdbcConnectionFactory - Connecting to jdbc:hsqldb:file:resources/org/apache/ctakes/dictionary/lookup/fast/sno_rx_16ab/sno_rx_16ab:
..14 May 2018 13:10:58  INFO ENGINE - open start - state not modified
........................
14 May 2018 13:11:06  INFO JdbcConnectionFactory -  Database connected
14 May 2018 13:11:06  INFO JdbcRareWordDictionary - Connected to cui and term table CUI_TERMS
14 May 2018 13:11:06  INFO JdbcConceptFactory - Connected to concept table TUI with class TUI
14 May 2018 13:11:06  INFO JdbcConceptFactory - Connected to concept table RXNORM with class LONG
14 May 2018 13:11:06  INFO JdbcConceptFactory - Connected to concept table PREFTERM with class PREFTERM
14 May 2018 13:11:06  INFO JdbcConceptFactory - Connected to concept table SNOMEDCT_US with class LONG
14 May 2018 13:11:06  INFO ContextAnnotator - Using left , right scope sizes: 10 , 10
14 May 2018 13:11:06  INFO ContextAnnotator - Using scope order: LEFT,RIGHT
14 May 2018 13:11:06  INFO ContextAnnotator - SCOPE ORDER: [1, 3]
14 May 2018 13:11:06  INFO ContextAnnotator - Using context analyzer: org.apache.ctakes.necontexts.status.StatusContextAnalyzer
14 May 2018 13:11:06  INFO StatusContextAnalyzer - initBoundaryData() called for ContextInitializer
14 May 2018 13:11:06  INFO ContextAnnotator - Using context consumer: org.apache.ctakes.necontexts.status.StatusContextHitConsumer
14 May 2018 13:11:06  INFO ContextAnnotator - Using lookup window type: org.apache.ctakes.typesystem.type.textspan.Sentence
14 May 2018 13:11:06  INFO ContextAnnotator - Using focus type: org.apache.ctakes.typesystem.type.textsem.IdentifiedAnnotation
14 May 2018 13:11:06  INFO ContextAnnotator - Using context type: org.apache.ctakes.typesystem.type.syntax.BaseToken
14 May 2018 13:11:06  INFO ContextAnnotator - Using left , right scope sizes: 7 , 7
14 May 2018 13:11:06  INFO ContextAnnotator - Using scope order: LEFT,RIGHT
14 May 2018 13:11:06  INFO ContextAnnotator - SCOPE ORDER: [1, 3]
14 May 2018 13:11:06  INFO ContextAnnotator - Using context analyzer: org.apache.ctakes.necontexts.negation.NegationContextAnalyzer
14 May 2018 13:11:06  INFO NegationContextAnalyzer - initBoundaryData() called for ContextInitializer
14 May 2018 13:11:06  INFO ContextAnnotator - Using context consumer: org.apache.ctakes.necontexts.negation.NegationContextHitConsumer
14 May 2018 13:11:06  INFO ContextAnnotator - Using lookup window type: org.apache.ctakes.typesystem.type.textspan.Sentence
14 May 2018 13:11:06  INFO ContextAnnotator - Using focus type: org.apache.ctakes.typesystem.type.textsem.IdentifiedAnnotation
14 May 2018 13:11:06  INFO ContextAnnotator - Using context type: org.apache.ctakes.typesystem.type.syntax.BaseToken
14 May 2018 13:11:06  INFO SentenceDetector - Sentence detector model file: org/apache/ctakes/core/sentdetect/sd-med-model.zip
14 May 2018 13:11:06  INFO POSTagger - POS tagger model file: org/apache/ctakes/postagger/models/mayo-pos.zip
14 May 2018 13:11:07  INFO LvgCmdApiResourceImpl - Loading NLM Norm and Lvg with config file = /Users/weng/git/clneg/src/ctakes/resources/org/apache/ctakes/lvg/data/config/lvg.properties
14 May 2018 13:11:07  INFO LvgCmdApiResourceImpl -   config file absolute path = /Users/weng/git/clneg/src/ctakes/resources/org/apache/ctakes/lvg/data/config/lvg.properties
14 May 2018 13:11:07  INFO LvgCmdApiResourceImpl - cwd = /Users/weng/git/clneg/src/ctakes
14 May 2018 13:11:07  INFO LvgCmdApiResourceImpl - cd /Users/weng/git/clneg/src/ctakes/resources/org/apache/ctakes/lvg/
14 May 2018 13:11:07  INFO ENGINE - open start - state not modified
14 May 2018 13:11:07  INFO ENGINE - dataFileCache open start
14 May 2018 13:11:07  INFO LvgCmdApiResourceImpl - cd /Users/weng/git/clneg/src/ctakes
14 May 2018 13:11:07  INFO DrugMentionAnnotator - Finite state machines loaded.
14 May 2018 13:11:10  INFO ClearNLPDependencyParserAE - using Morphy analysis? true
Loading configuration.
Loading feature templates.
Loading lexica.
Loading model:
........................................................................................
Loading configuration.
Loading feature templates.
Loading model:
.
Loading configuration.
Loading feature templates.
Loading lexica.
Loading model:
...
Loading configuration.
Loading feature templates.
Loading lexica.
Loading model:
................................
Loading model:
.............................
14 May 2018 13:11:25  INFO ConstituencyParser - Initializing parser...
Running CPE
To abort processing, type "abort" and press enter.
CPM Initialization Complete
14 May 2018 13:11:28  INFO SentenceDetector - Starting processing.
14 May 2018 13:11:28  INFO TokenizerAnnotatorPTB - process(JCas) in org.apache.ctakes.core.ae.TokenizerAnnotatorPTB
14 May 2018 13:11:28  INFO LvgAnnotator - process(JCas)
14 May 2018 13:11:28  INFO ContextDependentTokenizerAnnotator - process(JCas)
14 May 2018 13:11:28  INFO POSTagger - process(JCas)
14 May 2018 13:11:28  INFO Chunker -  process(JCas)
14 May 2018 13:11:28  INFO ChunkAdjuster -  process(JCas)
14 May 2018 13:11:28  INFO ChunkAdjuster -  process(JCas)
14 May 2018 13:11:28  INFO AbstractJCasTermAnnotator - Starting processing
14 May 2018 13:11:28  INFO AbstractJCasTermAnnotator - Finished processing
14 May 2018 13:11:28  INFO DrugMentionAnnotator - process(JCas)
14 May 2018 13:11:29  INFO MaxentParserWrapper - Started processing: tmp
14 May 2018 13:11:31  INFO MaxentParserWrapper - Done parsing: tmp
14 May 2018 13:11:31  INFO CasConsumer - Started
Completed 1 documents; 819 characters
Total Time Elapsed: 39441 ms
Initialization Time: 35566 ms
Processing Time: 3875 ms


 ------------------ PERFORMANCE REPORT ------------------

Component Name: lines from file collection reader
Event Type: Process
Duration: 228ms (5.93%)
Result: success
Component Name: AggregatePlaintextFastUMLSProcessor
Event Type: Analysis
Duration: 3494ms (90.82%)
Sub-events:
	Component Name: Chunker
	Event Type: Analysis
	Duration: 68ms (1.77%)

	Component Name: TokenizerAnnotatorPTB
	Event Type: Analysis
	Duration: 23ms (0.6%)

	Component Name: ContextDependentTokenizerAnnotator
	Event Type: Analysis
	Duration: 24ms (0.62%)

	Component Name: UmlsLookupAnnotator
	Event Type: Analysis
	Duration: 71ms (1.85%)

	Component Name: StatusAnnotator
	Event Type: Analysis
	Duration: 0ms (0%)

	Component Name: NegationAnnotator
	Event Type: Analysis
	Duration: 0ms (0%)

	Component Name: ExtractionPrepAnnotator
	Event Type: Analysis
	Duration: 6ms (0.16%)

	Component Name: Sentence Detector annotator
	Event Type: Analysis
	Duration: 16ms (0.42%)

	Component Name: Adjust NP in NP NP to span both
	Event Type: Analysis
	Duration: 6ms (0.16%)

	Component Name: Adjust NP in NP PP NP to span all three
	Event Type: Analysis
	Duration: 5ms (0.13%)

	Component Name: SimpleSegmentAnnotator
	Event Type: Analysis
	Duration: 12ms (0.31%)

	Component Name: POSTagger
	Event Type: Analysis
	Duration: 34ms (0.88%)

	Component Name: LVG Annotator
	Event Type: Analysis
	Duration: 275ms (7.15%)

	Component Name: DrugMentionAnnotator
	Event Type: Analysis
	Duration: 106ms (2.76%)

	Component Name: org.apache.ctakes.assertion.medfacts.cleartk.GenericCleartkAnalysisEngine
	Event Type: Analysis
	Duration: 151ms (3.93%)

	Component Name: org.apache.ctakes.assertion.medfacts.cleartk.HistoryCleartkAnalysisEngine
	Event Type: Analysis
	Duration: 136ms (3.54%)

	Component Name: org.apache.ctakes.assertion.medfacts.cleartk.PolarityCleartkAnalysisEngine
	Event Type: Analysis
	Duration: 123ms (3.2%)

	Component Name: org.apache.ctakes.assertion.medfacts.cleartk.SubjectCleartkAnalysisEngine
	Event Type: Analysis
	Duration: 67ms (1.74%)

	Component Name: org.apache.ctakes.assertion.medfacts.cleartk.UncertaintyCleartkAnalysisEngine
	Event Type: Analysis
	Duration: 64ms (1.66%)

	Component Name: ClearNLPDependencyParserAE
	Event Type: Analysis
	Duration: 249ms (6.47%)

	Component Name: ClearNLPSemanticRoleLabelerAE
	Event Type: Analysis
	Duration: 35ms (0.91%)

	Component Name: ConstituencyParserAnnotator
	Event Type: Analysis
	Duration: 2019ms (52.48%)

	Component Name: Fixed Flow Controller
	Event Type: Analysis
	Duration: 3ms (0.08%)

Component Name: AggregatePlaintextFastUMLSProcessor
Event Type: End of Batch
Duration: 1ms (0.03%)
Component Name: Write CAS to XML file
Event Type: Analysis
Duration: 124ms (3.22%)
Component Name: Write CAS to XML file
Event Type: End of Batch
Duration: 0ms (0%)

/Users/weng/git/clneg/src/concept_extraction.py:91: SettingWithCopyWarning:
A value is trying to be set on a copy of a slice from a DataFrame

See the caveats in the documentation: http://pandas.pydata.org/pandas-docs/stable/indexing.html#indexing-view-versus-copy
  df['negation'][idx] = 0
/Users/weng/git/clneg/src/concept_extraction.py:92: SettingWithCopyWarning:
A value is trying to be set on a copy of a slice from a DataFrame

See the caveats in the documentation: http://pandas.pydata.org/pandas-docs/stable/indexing.html#indexing-view-versus-copy
  df['sent_id'][idx] = i + 1 # sent_id from 1
/Users/weng/git/clneg/src/concept_extraction.py:93: SettingWithCopyWarning:
A value is trying to be set on a copy of a slice from a DataFrame

See the caveats in the documentation: http://pandas.pydata.org/pandas-docs/stable/indexing.html#indexing-view-versus-copy
  df['sent_loc'][idx] = int(df['start'][idx]) - nl[0] + 1 # sent_loc also start from 1

--- parse negated part of the sentence ---

8
8

--- Constituency tree parsing ---

sent: 0
original: chest x-ray is negative for infiltration .	 [NEGATED]

neg part: negative for infiltration .
negated term: negative for
--- tregex/tsurgeon with negated type: ADJP-A
constituency tree: (TOP (NN infiltration))
>> infiltration
>> negated span: (29, 40)

main.py:178: SettingWithCopyWarning:
A value is trying to be set on a copy of a slice from a DataFrame

See the caveats in the documentation: http://pandas.pydata.org/pandas-docs/stable/indexing.html#indexing-view-versus-copy
  df1['negation'][idx] = 1
sys:1: SettingWithCopyWarning:
A value is trying to be set on a copy of a slice from a DataFrame

See the caveats in the documentation: http://pandas.pydata.org/pandas-docs/stable/indexing.html#indexing-view-versus-copy
sent: 1
original: infection is ruled out .	 [NEGATED]

neg part: infection is ruled out
negated term: is ruled out
--- tregex/tsurgeon with negated type: ADJP-P
constituency tree: (TOP (NN infection))
>> infection
>> negated span: (1, 9)

sent: 2
original: the patient did not exhibit the sign of infection .	 [NEGATED]

neg part: not exhibit the sign of infection .
negated term: not exhibit
--- tregex/tsurgeon with negated type: ADVP-A
constituency tree: (TOP (VB exhibit) (NP (NP (DT the) (NN sign)) (PP (IN of) (NP (NN infection)))))
>> exhibit the sign of infection
>> negated span: (21, 49)

sent: 3
original: infection not seen .	 [NEGATED]

neg part: infection not seen
negated term: not seen
--- tregex/tsurgeon with negated type: ADVP-P
constituency tree: (TOP (S (NP (NP (NN infection)) (RB not)) (VP (VBN seen))))
>> infection not seen
>> negated span: (1, 18)

sent: 4
original: there is no significant congestive heart failure .	 [NEGATED]

neg part: no significant congestive heart failure .
negated term: no significant
--- tregex/tsurgeon with negated type: NP
constituency tree: (TOP (NP (NP (JJ significant) (JJ congestive) (NN heart)) (NN failure) (. .)))
>> significant congestive heart failure .
>> negated span: (13, 50)

sent: 5
original: the patient is free of malignancy .	 [NEGATED]

neg part: free of malignancy .
negated term: free of
--- tregex/tsurgeon with negated type: PP
constituency tree: (TOP (NN malignancy))
>> malignancy
>> negated span: (24, 33)

sent: 6
original: the examination can not see the tumor .	 [NEGATED]

neg part: not see the tumor .
negated term: not see
--- tregex/tsurgeon with negated type: ADVP-A
constituency tree: (TOP (VB see) (NP (DT the) (NN tumor)))
>> see the tumor
>> negated span: (25, 37)

sent: 7
original: renal malignancy was ruled out .	 [NEGATED]

neg part: renal malignancy was ruled out
negated term: was ruled out
--- tregex/tsurgeon with negated type: VP-P
constituency tree: (TOP (JJ renal) (NN malignancy))
>> renal malignancy
>> negated span: (1, 16)

main.py:187: SettingWithCopyWarning:
A value is trying to be set on a copy of a slice from a DataFrame.
Try using .loc[row_indexer,col_indexer] = value instead

See the caveats in the documentation: http://pandas.pydata.org/pandas-docs/stable/indexing.html#indexing-view-versus-copy
  df_s['start'] = df_s['start'].astype(int)
main.py:188: SettingWithCopyWarning:
A value is trying to be set on a copy of a slice from a DataFrame.
Try using .loc[row_indexer,col_indexer] = value instead

See the caveats in the documentation: http://pandas.pydata.org/pandas-docs/stable/indexing.html#indexing-view-versus-copy
  df_s['len'] = df_s['original'].str.len()

--- Final output ---

--- History of Present Illness ---
Infiltration(-), Communicable Diseases(-), Physical findings(-), Communicable Diseases(-), Communicable Diseases(-), Heart failure(-), Malignant Neoplasms(-), Neoplasms(-), Malignant Neoplasms(-)
\end{lstlisting}
\normalsize

\end{document}